# Artificial Intelligence can facilitate selfish decisions by altering the appearance of interaction partners

N. Köbis*, P. Lorenz-Spreen*, T. Ajaj, J.-F. Bonnefon, R. Hertwig, I. Rahwan^

*equal contributions

^ correspondence to: rahwan@mpib-berlin.mpg.de

**Abstract**

The ethical and psychological consequences of using Artificial Intelligence (AI) to manipulate our perception of others is an increasing phenomenon as image-altering filters proliferate on social media and video conferencing technologies. Here, we investigate the potential impact of a particular appearance-altering technology–blur filters–to investigate how individuals' behavior changes towards others. Our results consistently indicate an increase in selfish behavior at the expense of blurred individuals, suggesting blur filters can facilitate moral disengagement via depersonalization. These findings underscore the urgency for broader ethical discussions on AI technologies that alter our perception of others, encompassing transparency, consent, and the consequences of knowing that others can manipulate one's appearance. We highlight the potential role of *anticipatory* experiments in informing and developing responsible guidelines and policies ahead of technological reality.

## Introduction

Given the ongoing progress of Artificial Intelligence, we are approaching situations in which people use AI to mediate our perception of the environment, whether to facilitate antisocial behavior or to protect ourselves from emotional pain. AI is increasingly mediating communication between humans (Hancock et al., 2020; Jakesch et al., 2023; Purcell et al., 2023), and people have increasing control over the way they appear to others, as illustrated by the prevalence of beauty or youth-enhancing filters on platforms like Instagram or TikTok (Isakowitsch, 2023; Javornik et al., 2022; Ruggeri, 2023). Although research has been conducted on the motivations and consequences of altering one's own appearance using AI (Fribourg et al., 2021; Isakowitsch, 2023; Ryan-Mosley, 2022), there has been limited focus on the inverse behavior: AI manipulating how others appear to us.

There might be different motivations for altering the appearance of others. For example, some individuals may find it beneficial to sometimes reduce the emotional labor of their work. Content moderators, whose work entails watching disturbing stimuli for hours, already have access to filters that blur the materials they are looking at (Das et al., 2020; Steiger et al., 2021; Sullivan, 2019). In some cases, this is likely to be a controversial proposition, though: blurring people would be a form of depersonalization, and depersonalization is seen as an undesirable symptom of burnout in the helping professions, not a fix to prevent it (Rotenstein et al., 2018; West et al., 2018). Less controversially, some individuals may elect to blur the faces of others during an interaction if they want to stay blind to their appearance in order to make unbiased, impartial decisions. This is, in essence, a real-time hi-tech version of removing pictures from CVs (Behaghel et al., 2015) or conducting blind orchestra auditions (Goldin & Rouse, 2000). Hence, the emerging technology of real-time facial filters (in augmented reality) can provide new means for deliberate ignorance, i.e., choosing not to obtain potentially distressing information (Hertwig & Engel, 2016, 2021).

Blurring the face of one's opponent can be seen as functionally equivalent to an act of deliberate ignorance. It goes even further, as a person can take an active step to remove naturally available information, that is, the other's facial and thus personal identity. Blurring is thus enlisted in the service of regulating emotions and reaching two objectives: Maximization of self-interested allocation. Depersonalization makes this possible, as it reduces guilt when somebody acts in a selfish manner. Blurring can also be enlisted in the maximization of public welfare as opposed to individual welfare (i.e., charity game): Here, the blurring of the individual may make it more likely to allocate resources toward the public as opposed to the (depersonalized) individual.

In both cases, one can understand blurring as a way to increase the social and psychological distance between the decision maker (i.e., the dictator and the allocator) and the receiving party. The notion of psychological distance has been much studied in construal level theory(see e.g., Liberman & Trope, 2014); that of social distance by, for instance, Hofmann and colleagues (Hoffman et al., 1996).

On the other hand, blur filters and their depersonalizing effects may be used for more selfish means. We know that depersonalization leads to moral disengagement (Haslam & Loughnan, 2014): we care less about what happens to people we depersonalize, which makes it easier to behave selfishly at their expense or even engage in antisocial behavior toward them. Accordingly, some individuals may use blur filters to help them ignore the needs of others without paying the cost of guilt. In this article, we explore the dual use of blur filters through behavioral experiments in which participants see either the unblurred or blurred face of a partner before making a decision in the Dictator Game (i.e., split a pool of money between themselves and their partner) or in the Charity Game (i.e., split a pool of money between their partner and the World Food Programme). The Dictator Game captures situations in which a blur filter might distance people from their partner and thus facilitate selfish behavior, while the Charity Game captures situations in which a blur filter helps with a more impartial decision. We start with static situations in which participants only see a picture of their partner and then move on to interactive situations in which participants talk to their partner through a custom video conference software augmented with a real-time face-blurring filter. All data and code to reproduce the analyses and figures are openly available on the Open Science Framework via

https://osf.io/d8zpt/?view_only=b68257fe322f4003adf85d8d9c5994cb.

## Study 1

**Methods**

***Sample and Procedure.*** We recruited 208 participants ($M_{age}$= 35.43, $SD_{age}$= 11.12, 108 = female) via Prolific.co, out of which 200 completed both tasks. Participants first read the consent form informing them about the procedure, that their pictures would be deleted and not used for any analyses. Afterward, participants received illustrated instructions on the procedure (see OSF for screenshots and Figure 1 for an overview).

The instructions informed participants about how to provide a picture of themselves (i.e., upload existing or take a picture with their webcam) that had to adhere to some requirements (e.g., face needs to be visible, etc.). They also learned the games' structures and payoff implementation.

  ***Games.*** Participants were randomly assigned either to see the original unblurred version of their partners' pictures or their blurred version. This blurred version was built using a simple grid of uni-colored pixels, i.e., using a facial detection algorithm, we chopped the image into 10x10 pieces and applied the mean color to each of those tiles. Participants first played ten rounds of a Dictator Game where they decided how to split an endowment of $2 between themselves and a partner, who was another different participant in each round (see Figure 1, Panel A, left side). Importantly, they were informed that one of their decisions was payoff-relevant, and thus, real money would actually be paid out to the partner after the experiment. Next, participants played ten rounds of the Charity Game, where they decided how to split an endowment of $2 between the World Food Programme and their partner, who was different in each round (see Figure 1, Panel A, right side). In both games, participants made decisions on a slider that allowed them to choose any split, rounded to 10 cents. In both games, one of the ten rounds was randomly chosen for payment, and money was sent accordingly to the participants and the World Food Programme.

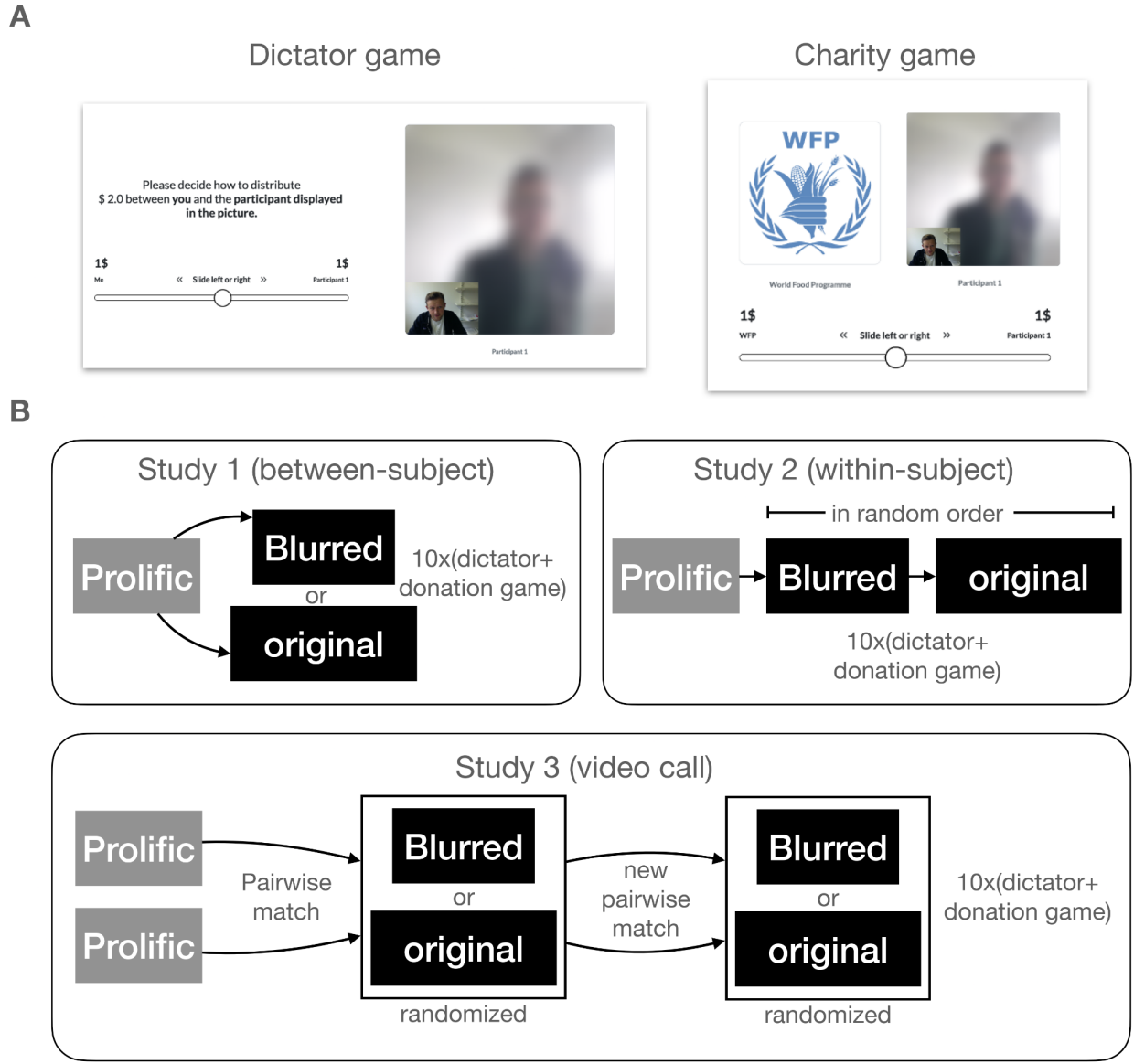

**Figure 1.** Panel A shows the implementation of the Dictator and the Charity Game with a blurred appearance of the recipient; Panel B shows the design of the middle panel shows the Charity Game with a blurred appearance of the recipient.

**Results**

    ***Dictator Game.*** We conducted a mixed-effects regression analysis to test whether participants allocated more money to themselves when the partner's image was blurred. Filter treatment was included as the binary predictor, together with random effects for the participant ID, the recipient ID, and the round number. The amount of money sent in

each Dictator Game served as the dependent variable. As shown in Figure 2 (Panel A), the results confirm the hypothesis that participants give less money to recipients when their appearance is blurred versus shown in its original ($B$ = -0.14, $SE$ = 0.05, $t(198.22)$ = -2.77, $p$ = .01).

*Charity Game.* We conducted the same mixed-effects regression analysis with the decisions in the Charity Game as the dependent variable. The results reveal no difference in Charity Game giving across the filter treatments ($B$ = -0.04, $SE$ = 0.06, $t(197.65)$ = 0.65, $p$ = .52), see also Figure 2, Panel B. Taken together, we find support for the notion that blur filters promote selfishness but not altruistic behavior towards a greater good in the form of donating to a charity.

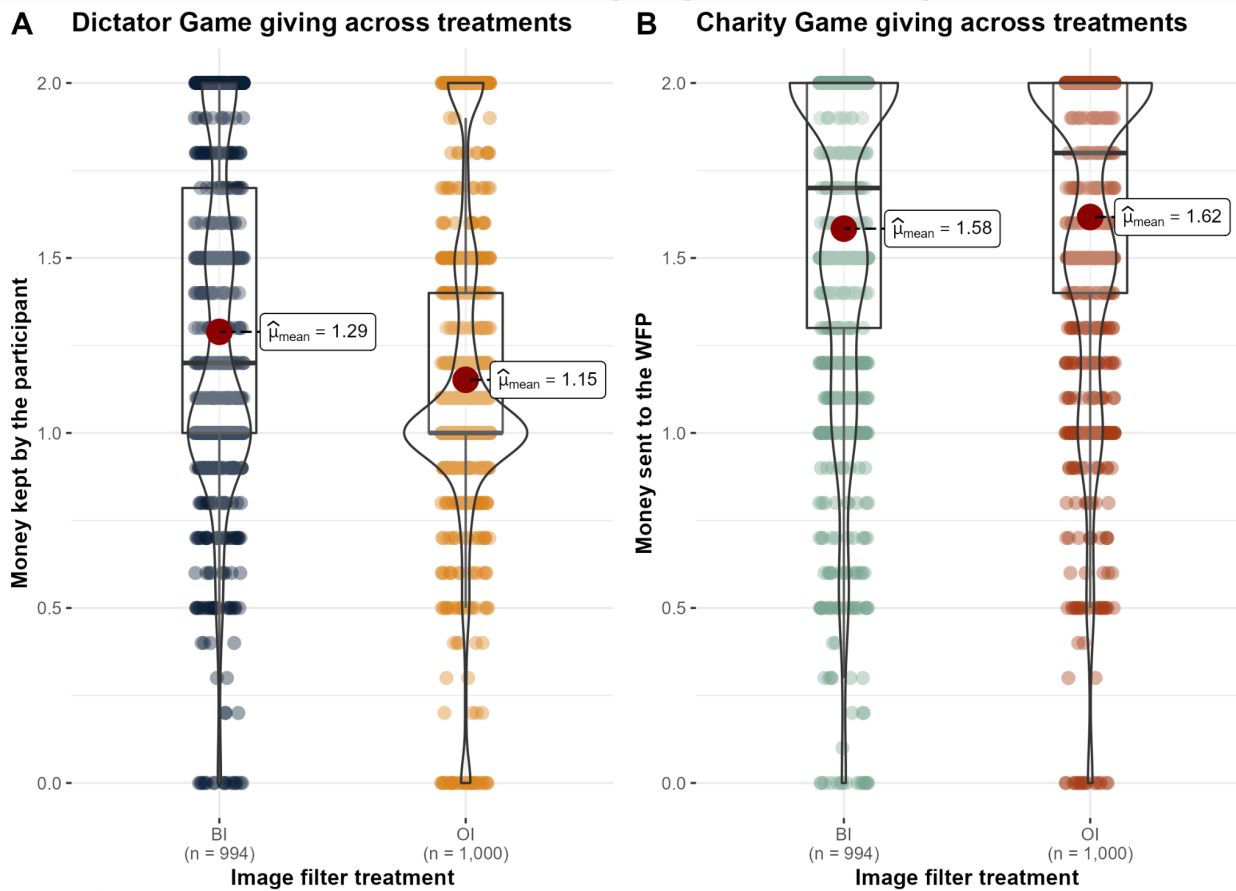

**Figure 2.** Violin plots depicting Dictator Game giving (Panel A) and Charity Game giving (Panel B) across ten rounds with the filter treatment applied on static images between participants. Dark red dots represent means. Boxes indicate the interquartile range; each dot shows a raw data point. BI refers to the treatment with a blurred image, and OI refers to the treatment with the original image. Plots were created with the Ggstatsplot package (Patil, 2021).

## Study 2

### Methods

We recruited 211 participants via Prolific.co and conducted a replication of Study 1, with only one change: Instead of manipulating the blur filter between-subject, the filter was manipulated within-subject, on a round-by-round basis.

**Results**

We again conducted mixed-effect regression analyses with the treatment as the predictor and random effects for participant ID, recipient ID, and round number. For the Dictator Game decisions, we replicate the findings of Study 1 (see Figure 3, Panel A). That is, participants give less money to partners shown with a blurry image than those shown with their original image ($B$ = -0.29, $SE$ = 0.02, $t(1980.69)$ = -18.52, $p < .001$). For Charity Game giving, we observe higher altruistic giving rates to charity when the partner is displayed with a blurry image ($B$ = -0.20, $SE$ = 0.01, $t(1974.35)$ = -14.43, $p < .001$), see also Figure 3 (Panel B). Thus, a within-subject replication consolidated the results observed in the Dictator Game and suggested a possible effect in the Charity Game in the expected direction, which was not observed with a between-subject manipulation.

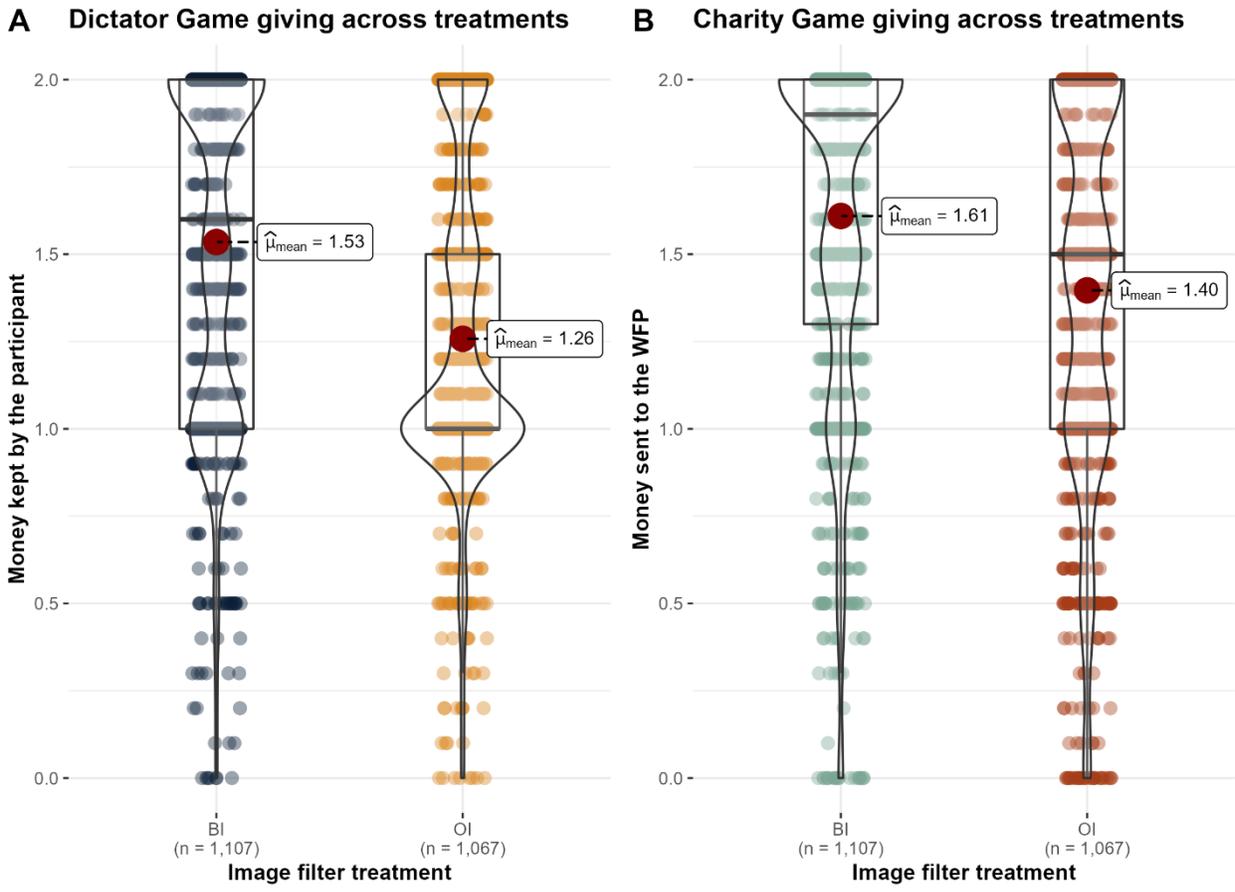

**Figure 3.** Violin plots depicting Dictator Game giving (Panel A) and Charity Game giving (Panel B) across ten rounds with the filter treatment applied on static images within subjects on a round-by-round level. Dark red dots represent means. Boxes indicate the interquartile range; each dot shows a raw data point. BI refers to the treatment with a blurred image, and OI refers to the treatment with the original image.

# Study 3

**Methods**

Study 3 involved 200 participants recruited via Prolific.co and used the same games as the previous studies. Instead of using static pictures of partners, we let participants enter a video conference call with their partners at the beginning of each round. The software for the video conference call was developed for the purpose of this research in order to maximize the privacy of participants and make it easy to run our own filters. Participants only played 5 rounds of each game due to the greater logistic complexity of real-time matching. At the beginning of each round, participants and their partners were asked to keep their webcam switched on, and we used the open-source software libfacedetection (Yu, 2021) to detect whether they actually displayed their faces. If so, they engaged in a 30 seconds video call. On a round-by-round level, the partner's face was randomly assigned to be left unblurred or blurred using a real-time filter that tracked their movement (and left unblurred everything that was not their face). We used the same form of blurring filter as in Study 1, just this time on the moving facial image.

**Results**

We conducted mixed-effects regression analyses using the filter treatment as a predictor and random effects for the participant ID, the respondent ID, and the round number. For the Dictator Game, we again replicated the previously observed effects. Again, participants gave less to recipients with a blurred (vs. original) appearance during the video call ($B$ = -0.17, $SE$ = 0.03, $t(685.54)$ = -6.69, $p$ < .001); see also Figure 4 (Panel A). For the Charity Game, we find the opposite effects observed in Study 2 (see Figure 4,

Panel B). Namely, participants gave less to the charity when the recipient's appearance was blurred (vs. in its original) (*B* = 0.23, *SE* = 0.03, *t*(668.82) = 7.88, *p* < .001).

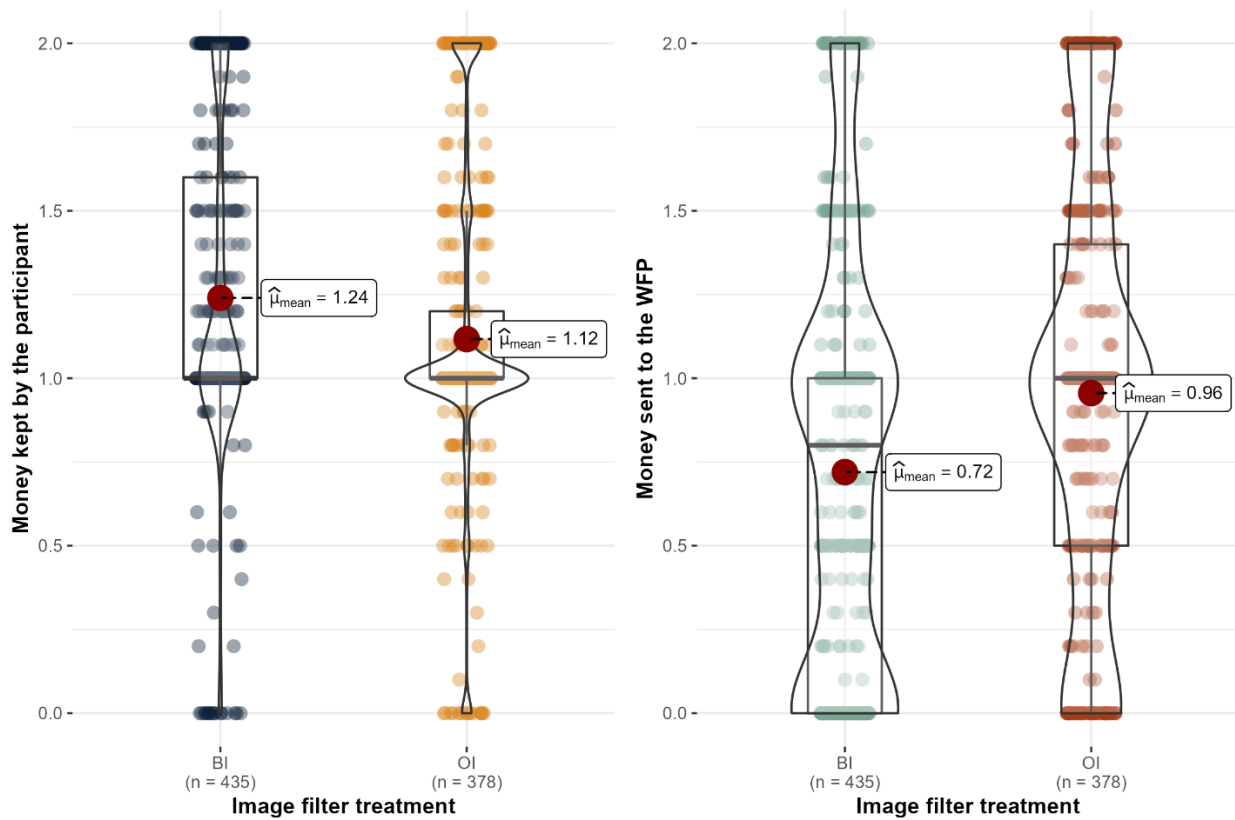

**Figure 4.** Violin plots depicting Dictator Game giving (Panel A) and Charity Game giving (Panel B) across 5 rounds. The filter treatment was applied in a real-time video conferencing interaction within-subjects on a round-by-round level. Dark red dots represent means. Boxes indicate the interquartile range; each dot shows a raw data point. BI refers to the treatment with a blurred image, and OI refers to the treatment with the original image.

### Mini Meta-Analysis

Figure 5 summarizes the effects of the blur filter in the Dictator Game and the Charity Game across the three studies, as well as the summary of the effect obtained through a mini meta-analysis aggregating the three studies. The blur filter has a robust effect in the Dictator Game across the three studies, increasing selfishness by approximately 15%. In contrast, the effect of the blur filter in the Charity Game is very inconsistent, and its summary effect is accordingly uninformative. We have no convincing explanation for this inconsistency across studies.

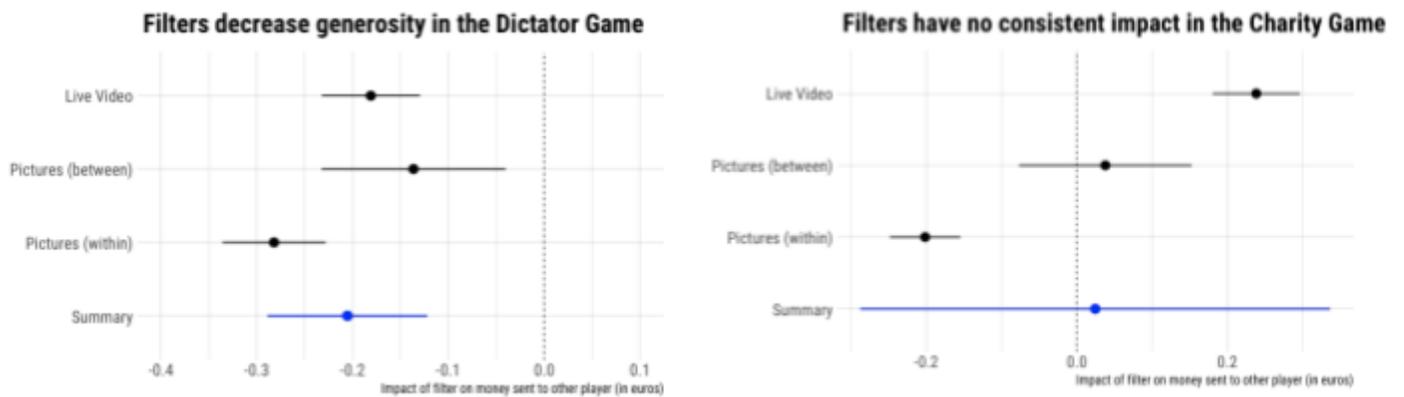

**Figure 5**. Results of a mini meta-analysis aggregating the results of the three studies.

## Discussion

In the TV series Black Mirror episode *Men Against Fire,* soldiers perceive the world through a real-time AI filter that turns their adversaries into monstrous mutants to overcome their reluctance to kill. These science fiction scenarios explore the ethical and psychological implications of face-blurring technology. Given the widespread use of AI filters, we are already familiar with the fact that the people we see online may look different than their physical selves. What we do not fully realize yet is that filters could also change the way others see *us*. In the future, metaverse interactions and offline interactions mediated by augmented reality devices will offer endless possibilities for people to alter the way they see their environment or the way they see other people. In this work, we focused on a simple alteration, which is already easily implementable in real-time: blurring the faces of others.

Across three experiments, including an experiment involving real-time interactions via video conference, we observed a robust effect: people who use the blur filter behave more selfishly at the expense of the individuals whose faces are blurred. This result aligns with the idea that blur filters enable moral disengagement by depersonalizing the individuals we interact with. It informs the growing research that examines how interactions with and through AI shape social and ethical behavior (Köbis et al., 2021). We already knew that depersonalization could have such undesirable effects, but our research suggests that depersonalization will soon become an easy, accessible option afforded by technological devices. This development calls for a broader discussion on managing the availability and use of AI technologies that change how others appear to us.

On the other hand, our results do not draw a clear picture of the effect of blurring on the behavior in the donation game. We believe the conflicting results between the static- and the video-format experiments could have their roots in two aspects. The choice between a charity or another individual is less morally unbalanced and clear than for the Dictator Game, as both options result in an altruistic act in principle. The social interaction with the partner is more pronounced in the video format through the interactive and simultaneous character and could potentially explain the flipped effect. (1) Even the interaction with a blurred partner is more influential than donating to an anonymous charity in study 3, while a static, blurred image in studies 1 and 2 does not have that impact. (2) Relatedly, concerns of reciprocity could play a role, as participants in the video-format experiment were aware that they were at the moment in the respective role for the partner and hence were more generous towards them than in the static format.

For the small effects in the static format of studies 1 and 2, we believe that ceiling effects could be a potential explanation as the preference towards the charity instead of the static image of the partner and hence the effects of the blurring were not recognizable.

The limitations of our studies include the ecological validity of the two games we included, as filters will play a role in various social situations and decisions. Future work should also investigate how in- vs. out-group partners might change the effects of blurring filters as well as if and when people decide to apply filters to others if they are given a choice.

The ongoing ethical conversation about the use of AI filters has primarily focused on the modification of the user's own appearance, for example, its relation to self-esteem and interest in cosmetic surgery (Chen et al., 2019). This conversation must be extended to include the use of filters that we willfully engage to change the way others appear to us. Consider, for example, the issue of transparency and consent. Should individuals have the right to know if their appearance is being altered by someone else's filter, how would this information be communicated, and should the use of the filter require explicit consent from the person whose appearance is altered?

In the case of blur filters, a case could be made that people should have agency about the way they appear to others, or at least be informed of the way they appear to others, given that blur filters may lead to depersonalization, moral disengagement, and selfish behavior – and these considerations may extend to other filters, for example, filters that alter the appearance of people in a way they find offensive. This discussion will be difficult, though, given the current absence of social or legal norms around this use of technology and the complexity of its ethical implications. For example, we may argue that people have a right to know how they appear to others since not knowing would violate their personal autonomy. Yet, on the flip side are the psychological costs of knowing that one is depersonalized by others, such as stress, low self-esteem, and feelings of powerlessness.

In conclusion, the fact that augmented reality filters can be used to depersonalize others and promote selfish behavior has significant implications for both individuals and society at large. Expanding the ethical conversation around AI filters, addressing issues of consent, and understanding the psychological impact of technological depersonalization are important steps in navigating this emerging landscape. The academic community must work collaboratively to help develop responsible guidelines and policies that protect well-being and autonomy while acknowledging the complexity of the ineluctable future in which everyone has control over the way others appear to them.